\documentclass[conference]{IEEEtran}
\IEEEoverridecommandlockouts

\usepackage{cite}
\usepackage{amsmath,amssymb,amsfonts}
\usepackage{algorithmic}
\usepackage{graphicx}
\usepackage{textcomp}
\usepackage{xcolor}
\usepackage{graphicx}
\usepackage{subcaption}
\usepackage[font=small]{caption}
\usepackage{algorithm}
\usepackage{algorithmic}
\usepackage{subcaption}
\usepackage{amsmath,algorithm}
\usepackage{mathrsfs}
\usepackage{subcaption}  
\usepackage{graphicx}    
\usepackage{xcolor}
\usepackage{booktabs} 
\usepackage{tabu}
\usepackage{tabularray}
\usepackage{soul}
\def\BibTeX{{\rm B\kern-.05em{\sc i\kern-.025em b}\kern-.08em
    T\kern-.1667em\lower.7ex\hbox{E}\kern-.125emX}}
\begin{document}

\title{Toward a Low-Cost Perception System in Autonomous Vehicles: A Spectrum Learning Approach\\
}

\author{
    Mohammed Alsakabi\textsuperscript{\rm 1},
    Aidan Erickson\textsuperscript{\rm 1},
    John M. Dolan\textsuperscript{\rm 2},
    Ozan K. Tonguz\textsuperscript{\rm 1} \\
    
    \textsuperscript{\rm 1}Department of Electrical and Computer Engineering, College of Engineering \\
    \textsuperscript{\rm 2}The Robotics Institute, School of Computer Science \\
    Carnegie Mellon University, Pittsburgh, PA, United States\\
    \{malsakabi, aerickson\}@cmu.edu , \{jdolan, tonguz\}@andrew.cmu.edu
}

\maketitle
\begin{abstract}

We present a cost-effective new approach for generating denser depth maps for Autonomous Driving (AD) and Autonomous Vehicles (AVs) by integrating the images obtained from deep neural network (DNN) 4D radar detectors with conventional camera RGB images. Our approach introduces a novel pixel positional encoding algorithm inspired by Bartlett's spatial spectrum estimation technique. This algorithm transforms both radar depth maps and RGB images into a unified pixel image subspace called the Spatial Spectrum, facilitating effective learning based on their similarities and differences. This method effectively leverages high-resolution camera images to train radar depth map generative models, addressing the limitations of conventional radar detectors in complex vehicular environments, thus sharpening the radar output. We develop spectrum estimation algorithms tailored for radar depth maps and RGB images, a comprehensive training framework for data-driven generative models, and a camera-radar deployment scheme for AV operation. Our results demonstrate that our approach also outperforms the state-of-the-art (SOTA) by 24.24\% 52.59\% in terms of the Unidirectional Chamfer Distance (UCD) and the Mean Absolute Error (MAE), respectively. Python codes and demonstration videos are available on our GitHub repository\footnote{https://shorturl.at/FkdJC}.
\end{abstract}

\section{Introduction}

\par Autonomous Vehicles (AV) employ various sensors for comprehensive navigation and environmental perception, each contributing distinct advantages and limitations \cite{wang2019multi}. RGB cameras are attractive due to their affordability and high accuracy in optimal lighting conditions, but they struggle in low-visibility scenarios like darkness, heavy rain, or fog, which impede obstacle detection. Lidars offer precise 3D mapping and depth measurements, yet they consume high power, are susceptible to adverse weather conditions, and sometimes require mechanical parts for rotational scanning, raising the overall manufacturing and operational costs. Radars, however, stand out for their robustness in adverse weather, cost-effectiveness, low power consumption, and ability to offer a theoretical angular resolution comparable to lidars. 

\begin{figure}[t!]
  \centering
  \includegraphics[width=\linewidth]{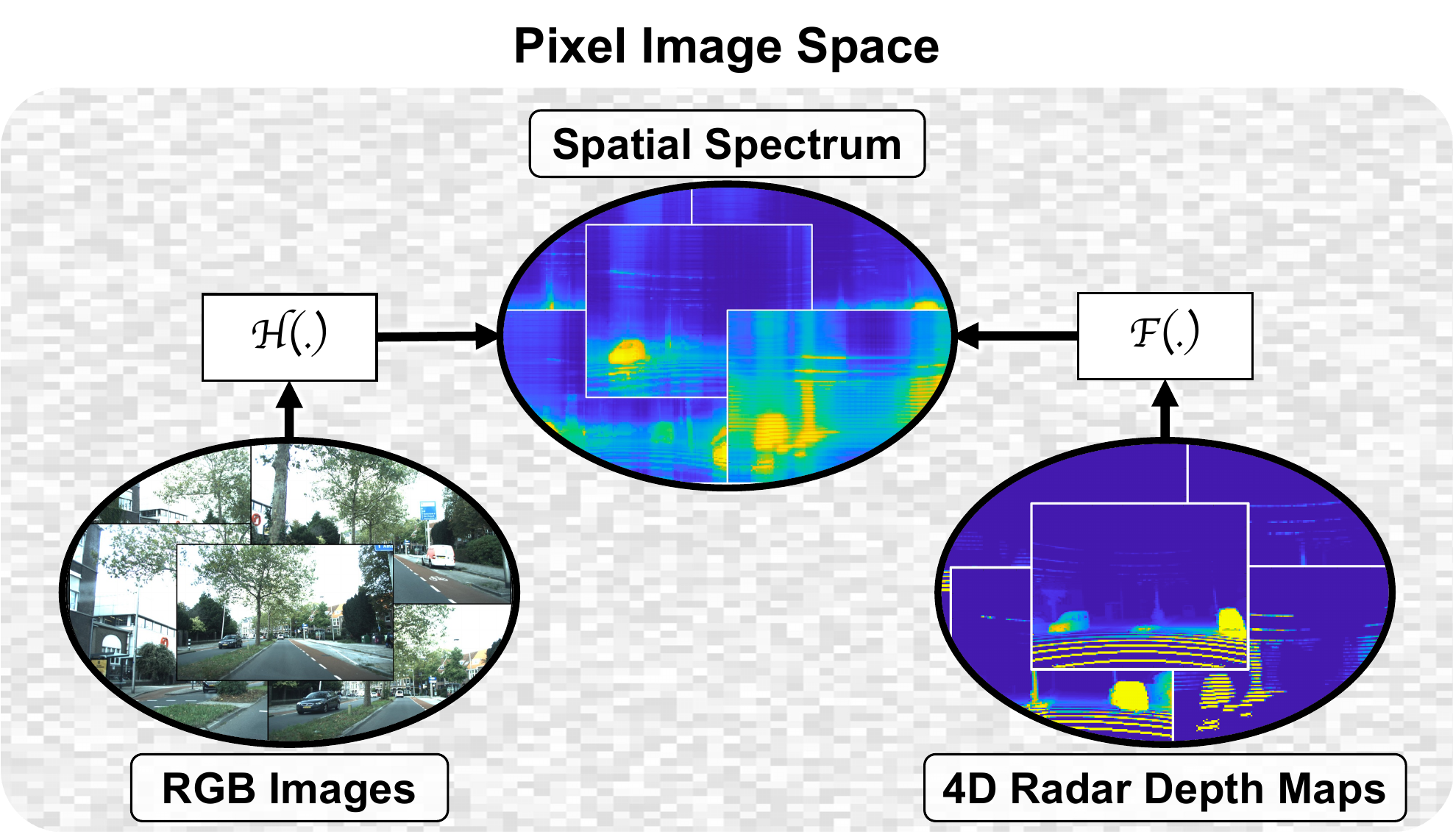}
  \caption{Conceptual figure depicting the proposed spectrum-based transformation. The 'RGB' and 'Depth Maps' subspaces represent natural images captured by conventional cameras and the windshield view depth maps that lidars and radars produce, respectively. The 'Spatial Spectrum' subspace includes the special frequency 2D spectrum of bases that constitute images in the other subspaces.}
  \label{fig: image space}
  \vspace{-22px} 
\end{figure}

\par Automotive radars measure range, azimuth, and velocity, with 4D radars adding elevation to their traditional measurements. Unlike 3D radars, 4D radars are capable of estimating object heights without using speculative models. For such radars, extracting precise angular information involves a two-stage process: spatial spectrum estimations followed by Constant False Alarm Rate (CFAR) detectors like cell-averaging CFAR (CA-CFAR) and order-statistic CFAR (OS-CFAR) \cite{richards2010principles}. However, these conventional methods struggle in complex vehicular settings, producing sparse point clouds that limit accurate environmental representation \cite{khan2022comprehensive}. To address these limitations, data-driven approaches using deep neural networks (DNNs) have been reported \cite{brodeski2019deep, cheng2022novel, roldan2024see}. For instance, \cite{roldan2024see} employs a ResNet18 network \cite{chen2017rethinking} trained on dense lidar point clouds, generating denser radar point clouds that more accurately represent object shapes and sizes.

\par In this paper, we introduce a data-driven approach to generate radar depth maps by integrating radar point clouds with camera images. Leveraging the similar field-of-view (FoV) between radar and camera modalities, employing a non-linear frequency pixel positional encoding algorithm and Bartlett's spatial spectrum estimation \cite{Bartlett1948} transforms radar depth maps and camera RGB images into a shared spatial spectrum subspace, via $\mathscr{F}(.)$ and $\mathscr{H}(.)$, as shown in Figure \ref{fig: image space}. This transformation can resolve the differences between the 4D radar depth maps and camera images by representing them as spectrums of a common set of basis functions, thus enabling spectrum-based learning \cite{gao2019deep}. Our approach allows high-resolution camera data to effectively supervise the generation of radar depth maps during offline training. After this offline training, the 4D radar model can operate independently of the camera, retaining radar advantages while producing sharper and denser depth maps essential for downstream tasks such as perception, tracking and rendering. Our contributions can be summarized as:

\begin{itemize}
\item We propose a pixel positional encoding algorithm that helps resolve the differences between the 4D radar and camera modalities, thus enabling spectrum-based learning for 4D radar depth maps.


\item We present experimental results for high-resolution spectrum estimations and depth map generations. Our approach produces sharper depth maps and significantly outperforms the state-of-the-art (SOTA), resulting in a reduction of 24.24\%, 18.52\%, 52.59\%, 10.41\% in Mean Absolute Error (MAE), Relative Absolute Error (REL), Unidirectional Chamfer Distance (UCD), and Bidirectional Chamfer Distance (BCD), respectively, which is quite significant. We also show that the estimated spectrum of camera and radar images results in an increase in the Pearson correlation and mutual information between the two modalities by factors of 3.88 and 76.69, respectively, indicating enhanced cross-modal alignment.

\end{itemize}

\section{Related Work}


\subsection{Datasets}
This work requires 4D radar data that include raw measurements in order to apply novel DNN detectors. Here, we review available 4D radar datasets.

The RaDelft, VoD, and K-Radar datasets are critical resources for automotive radar research, each offering unique capabilities. The RaDelft dataset provides synchronized 4D radar, lidar, camera, and odometry data from Delft, Netherlands, and has been used in deep learning to replicate lidar-like expressiveness with radar input \cite{roldan2024see}. Similarly, the VoD dataset integrates camera, BEV S3 lidar, 4D radar, and odometry data, offering annotations and bounding boxes for object detection, making it ideal for multimodal research and downstream tasks \cite{VoD}. The K-Radar dataset serves as a benchmark for object detection and classification with pre-processed radar data like range Doppler and range-angle maps, along with ground truth annotations. However, it lacks raw 4D radar signals, limiting its application in workflows requiring unprocessed radar data \cite{kradar}. In this work, we are using the RaDelft DL detector dataset, as VoD and K-radar do not provide raw radar signals that are sufficient for applying the detector developed in \cite{roldan2024see}.

\subsection{Multimodal Data Fusion for Depth Map Generation}

Several advanced methods leverage radar and camera fusion to enhance autonomous driving perception tasks. For example, \cite{long2021radar} introduces a two-stage pixel-level fusion method for depth completion, where radar-to-pixel association in the first stage maps radar returns to pixels using lidar-generated ground truth, and the second stage fuses this output with camera images featuring projected 3D radar point clouds. Similarly, \cite{singh2023depth} adopts a two-stage approach that uses RadarNet to generate a confidence map from RGB images and noisy 3D radar point clouds in the first stage, followed by depth map enhancement using the confidence map and RGB images. In contrast, \cite{li2024radarcam} proposes a four-stage framework for dense depth estimation that enhances monocular depth prediction with accurate metric scale information derived from sparse 3D radar data, using global scale alignment, transformer-based quasi-dense scale estimation, and local refinement. Expanding on these concepts, \cite{li2024semantic} develops a semantic-guided depth completion framework that incorporates a Radar Point Correction Module (RPCM) for radar refinement, a Semantic Divide-and-Conquer Module (SDCM) for category-specific tasks, and a Region Depth Diffusion Module (RDDM) for further depth refinement. Additionally, \cite{radarocc} targets 3D occupancy prediction by directly processing radar tensors with Doppler bin descriptors and spherical-based feature encoding, excelling in adverse weather conditions on the K-Radar dataset. 
\par Unlike the aforementioned methods, our approach shows that using the DNN detector model in [5] as a module, we can improve the depth maps generated by further training this model with images generated by cheap cameras. This paves the way for eliminating the use of lidars for real time perception in AVs.

\begin{figure*}[!t]
  \centering
  \begin{minipage}{\linewidth}
    \centering
    \begin{subfigure}[b]{\textwidth}
      \centering
      \includegraphics[width=0.83\linewidth]{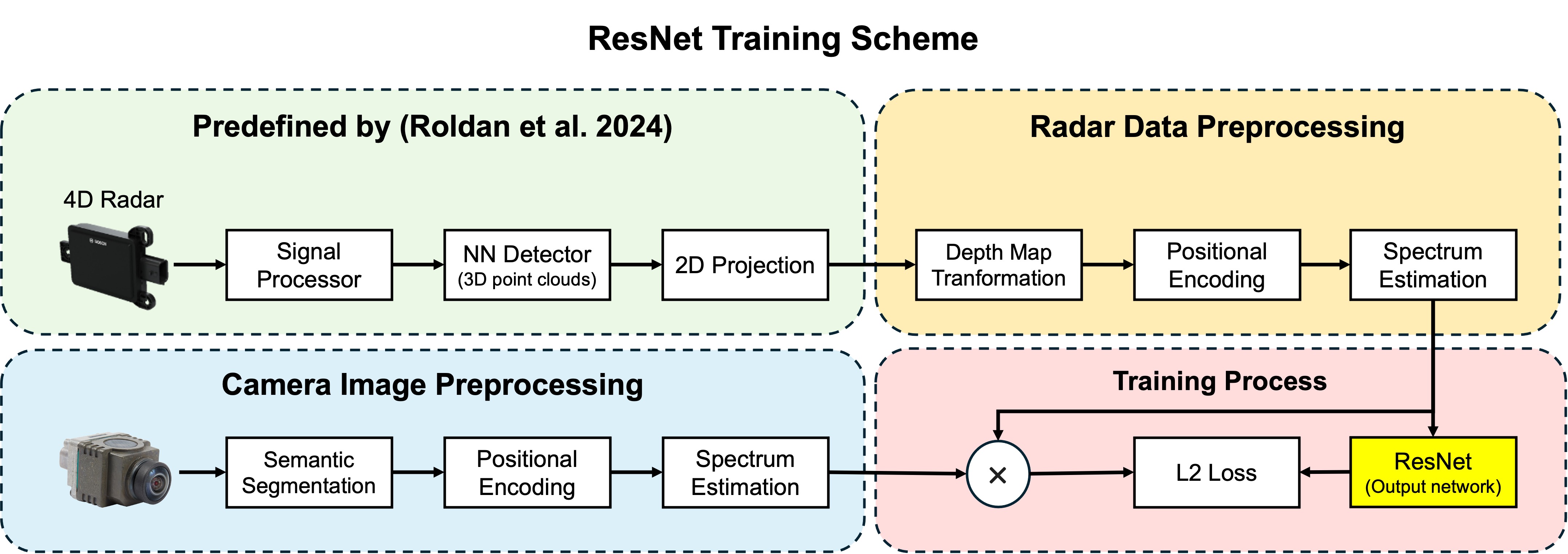}
      \caption{}
      \label{fig:RSUencountered1}
    \end{subfigure}
    \vspace{1em} 
    \begin{subfigure}[b]{\textwidth}
      \centering
      \includegraphics[width=0.83\linewidth]{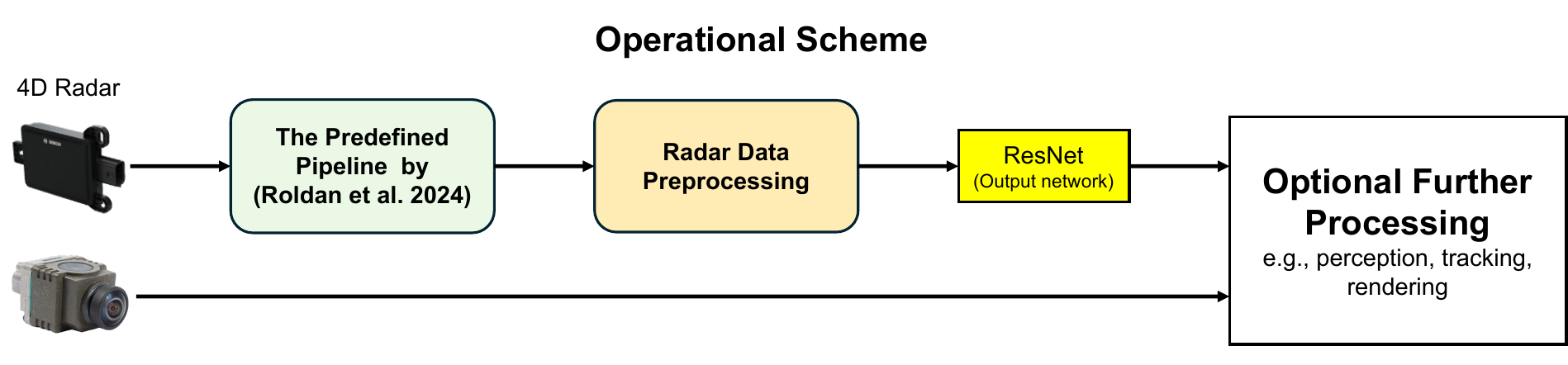}
      \caption{}
      \label{fig:RSUencountered2}
      \vspace{-11px}
    \end{subfigure}
  \end{minipage}
  \caption{The pipeline for our proposed method: (a) The offline network training scheme is divided into four modules: the predefined radar signal processing and detection (point-cloud extractor) by \cite{roldan2024see}, the preprocessing of camera and radar data defined under 'Proposed Approach', and the training process which takes the 'Radar Data Preprocessing' output as input and the filtered 'Camera Image Preprocessing' output as the ground truth for training. (b) The proposed operational deployment of the trained network in (a) considering the required lower-level processes, where radar and camera work independently of each other and provide data for any required further process.}
  \label{our_pipeline}
  \vspace{-11px}
\end{figure*}
\section{Background}
\subsection{Radar Detectors}


Radar detectors process signals to differentiate targets from noise by applying a decision threshold, comparing the signal strength to a predefined value, and outputting results as point clouds. Conventional radar detectors, like CA-CFAR and OS-CFAR, aim to maintain a consistent false alarm rate by dynamically adjusting decision thresholds based on surrounding noise and clutter. CA-CFAR works well in homogeneous environments but struggles with the heterogeneity of vehicular surroundings. OS-CFAR handles heterogeneous environments but requires precise prior knowledge of target numbers, which might be challenging to estimate. However, a recent deep learning-based radar detector proposed by \cite{roldan2024see}, trained on lidar point clouds, addresses these limitations but still struggles to capture dense scene representations compared to lidar. In this work, we leverage these sparse outputs to generate denser and more accurate depth maps. 

\subsection{Bartlett's Algorithm for Spatial Power Spectrum Estimation}
Bartlett's algorithm, also known as periodogram averaging \cite{Bartlett1948}, is used in signal processing and time series analysis to estimate the power spectral density of a random sequence. It divides the sequence into \( M \) non-overlapping segments, computes their periodograms, and averages them, with the number of segments being proportional to the spectral resolution. For received signals at spatially distant receptors, like signals received at different camera physical pixels or signals received in multi-antenna wireless systems, the $M$ segments correspond to the signals received at the $M$ receptor. The time difference between segments introduces a phase shift proportional to the spatial frequency \(\omega\), referenced against the complex sinusoidal signal at the first receptor\cite{AoA}, which is described as:

\begin{equation}
x_1(n) = e^{-j\omega n} , n=0,1,...,N
\end{equation}
where $N$ is the number of samples in the signal. Assuming that we have $M$ segments, each segment has a time delay that is translated into a phase shift in $x_1(n)$, expressed as:
\begin{equation}
x_m(n) = x_1(n)e^{-jm\phi} =e^{-j(\omega n + m\phi)} , m=0,1,...,M-1
\end{equation}
Hence, the $M$-segment matrix $\textbf{S}\in \mathbb{C}^{N\times M}$ is found as:
\begin{equation}
    \textbf{S} = \begin{bmatrix} \textbf{x}_{1} & \textbf{x}_{2} & \cdots & \textbf{x}_{M}  \end{bmatrix}
\end{equation}
Since we are computing the spatial power spectrum, our goal is to calculate the spectrum of the signal that is described by relative phases between segments, which can be found in the covariance matrix $\textbf{C}\in \mathbb{C}^{M\times M}$ of $\textbf{S}$ as:
\begin{equation}
    \textbf{C}= \frac{1}{N}\textbf{S}^H\textbf{S}
\end{equation}
where $H$ denotes the Hermitian or complex conjugate. Therefore, the power spectral density at $\phi$, $P(\phi)$, is found as:
\begin{equation}
    P(\phi) = \textbf{a}(\phi)^H\textbf{C}\textbf{a}(\phi)
\end{equation}
where $\textbf{a}(\phi) = \begin{bmatrix} 1 & e^{-j\phi} & e^{-j2\phi} & \cdots & e^{-j(M-1)\phi}  \end{bmatrix}^T$ is our basis vector. Note that the choice of basis vector depends on the application, pattern of interest, and desired resolution \cite{priestley1981spectral}. A different basis vector is used for our proposed approach and defined in the next section.

\begin{figure*}[t]
  \centering
  \includegraphics[width=0.92\linewidth]{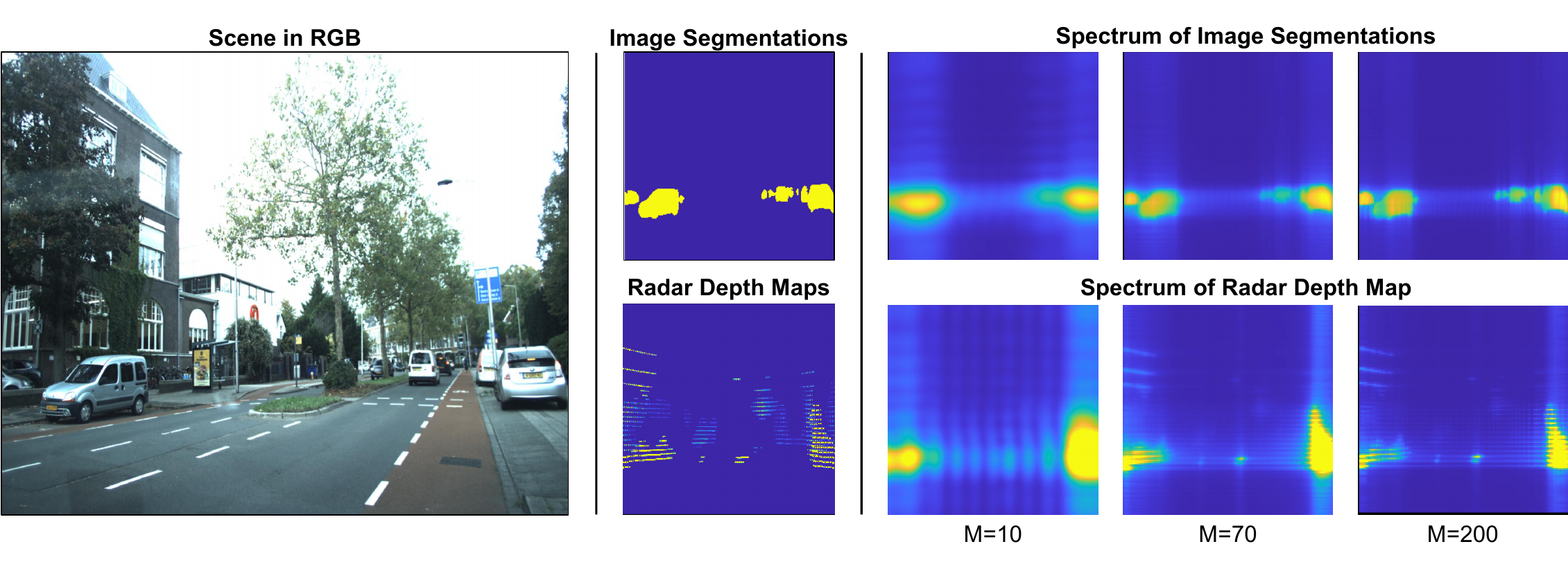}
  \caption{Results for $M=10, 70$ and $200$, and $\Phi=\Theta=(-70,70)$ against the original RGB scene on the left.}
  \label{fig:algorithm_results}
  \vspace{-11px}
  \centering
\end{figure*}

\section{Proposed Approach}
Unlike lidars, 4D imaging radars used in AV suffer from sparse scene representations. Our goal in this work is to bypass lidars and produce sharp 4D radar depth maps by passing the radar output to a data-driven depth map generator. Since the camera RGB images have different characteristics from radar depth maps (i.e., they come from different pixel image subspaces), we propose to compute their unified, constitutive basis vectors and transform them into their spatial spectrum representations using Bartlett’s algorithm. This bridges the gap between their original characteristics. With non-linear frequency progression basis vectors, we propose to encode the semantic segmentations of camera images and their corresponding radar depth maps to estimate their spatial spectrum. Interestingly, the spatial spectrum of both images includes additional frequency components proportional to $M$ caused by spectral leakage. These frequency components complete the depth for sparse point clouds, as well as introduce frequency bias that helps with fitting highly oscillatory data (the sharp camera images), as shown in Figure \ref{fig:algorithm_results} \cite{xu2024overview}. Figure \ref{fig:RSUencountered1} depicts this transformation and network training pipeline, while \ref{fig:RSUencountered2} shows the deployment pipeline for real-time operation, noting that the 4D radar and camera work independently but synchronously.


\subsection{Pixel Positional Encoding and Spectrum Estimation}
This encoding method aims to facilitate the transformation of radar and camera images into the spatial spectrum. Fast implementation of this encoding process starts with an initialization of a non-linear phase progression, $M$ segments' complex sinusoidal basis functions for horizontal and vertical axes, $\phi$ and $\theta$. Our basis functions differ from the standard Fourier basis functions, since we require higher resolution for the output \cite{pillai2012array, buckley1990spatial}. For that, we change the phase progression across basis functions from standard linear to non-linear, resulting in changing frequency across receptors. This changing frequency leads to higher-resolution spectrum \cite{richards2010principles}. Our segments are described as:
\begin{equation}
    x(m,\phi_n)=e^{-j\pi m sin(\phi_n)} , x(m,\theta_k)=e^{-j\pi m sin(\theta_k)}
    \label{eq:horizontal_basis}
\end{equation}
where $m$ is the segment index, noting that $M$ is proportional to spectrum resolution, and $\phi_n$ and $\theta_k \in \Phi$ and $\Theta$ are variation angles from the set $(-90, 90)$ with lengths $N$ and $K$, respectively. The covariance matrices of every $\phi$ and $\theta$ are defined as $\textbf{C}(\Phi)$ and $\textbf{C}(\Theta) \in \mathbb{C}^{N \times N}$ and $\mathbb{C}^{K \times K}$, in which each of their rows represents the periodograms $\textbf{y}(\phi_n)$ and $\textbf{y}(\theta_k)$. The joint 2D periodogram is:
\begin{equation}
    \textbf{Y}(\phi_n, \theta_k) = \textbf{y}(\theta_n)^T\textbf{y}(\phi_k)
\end{equation}
To calculate the final 2D spatial power spectrum $\textbf{P} \in \mathbb{R}^{N \times K}$ for an input image $\textbf{I}$, we iteratively encode all pixels of $\textbf{I}$ and calculate $P(n,k)$ as:
\begin{equation}
    P(n,k)= \sum_{n=0}^{N-1} \sum_{k=0}^{K-1} \left| \textbf{Y}(\phi_n, \theta_k) \circ \textbf{I}\right|
\end{equation}
where $\circ$ denotes an element-wise multiplication. Experimental results are presented in the next section.





\subsection{Radar Data Preprocessing}
This module conditions the DNN detector's depth map, $\textbf{I}_{radar}$, in the predecessor radar pipeline. The objective is to transform 2D depth maps into a spatial spectrum representation of its constitutive bases, $\textbf{P}_{radar}$, to satisfy the input characteristics in the following module, the 'Training Process'. This process of spatial spectrum estimation is defined as $\mathscr{F}(\textbf{I}, M)$, noting that $M$ is proportional to the output resolution.
\begin{equation}
    \textbf{P}_{radar}=\mathscr{F}(\textbf{I}_{radar}, M_{radar})
\end{equation}


\subsection{Camera Image Preprocessing}
In order to obtain the spatial spectrum representations for objects of interest, we transform the RGB image, $\textbf{I}_{cam}$, into its semantic segmentations, $\textbf{Seg}$, considering that this is highly dependent on the semantic segmentation accuracy and classes of the model in use. We use Deeplab v3 \cite{yurtkulu2019semantic} with ResNet101 \cite{chen2017rethinking} trained on the Cityscapes autonomous driving benchmark \cite{cordts2016cityscapes}. We then transform the semantic image into its spatial spectrum representation, $\textbf{P}_{cam}$, via $\mathscr{H}(\textbf{I}, M)$, which is the process of estimating the spatial spectrum of camera images.
\begin{equation}
    \textbf{P}_{cam}=\mathscr{H}(\textbf{I}_{cam}, M_{cam})=\mathscr{F}(\textbf{Seg}, M_{cam})
\end{equation}
Note that we require $M_{cam} > M_{radar}$ so that the radar spectrum images have a lower resolution, which leaves room for enhancement with deep learning models. 


\subsection{Network Training Process}
This module focuses on training a generative model that produces a denser and contour-accurate version of $\textbf{P}_{radar}$. As there are some objects captured by the semantic segmentation model that are not detectable by radar, and vice versa, element-wise multiplication produces the mutuality between both spectrum images which, thereafter, is fed into the learning as ground truth for training a ResNet, being optimized to reduce the difference through L2 loss. The process is described as:
\begin{equation}
    \textbf{P}_{radar} \circ \textbf{P}_{cam} = \text{ResNet}(\textbf{P}_{radar})
\end{equation}

\begin{figure*}[!t]
	\centering
	\begin{subfigure}{0.30\linewidth}
		\includegraphics[width=\linewidth]{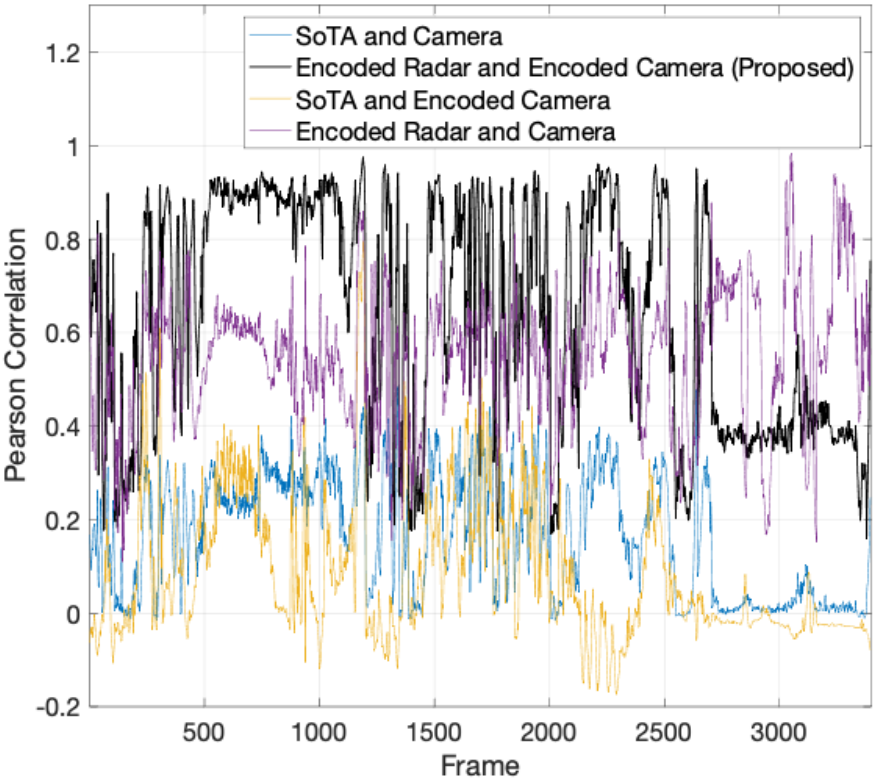}
		\caption{}
		\label{fig:Correlation}
	\end{subfigure}
        \hspace{0.01\textwidth}
	\begin{subfigure}{0.30\linewidth}
		\includegraphics[width=\linewidth]{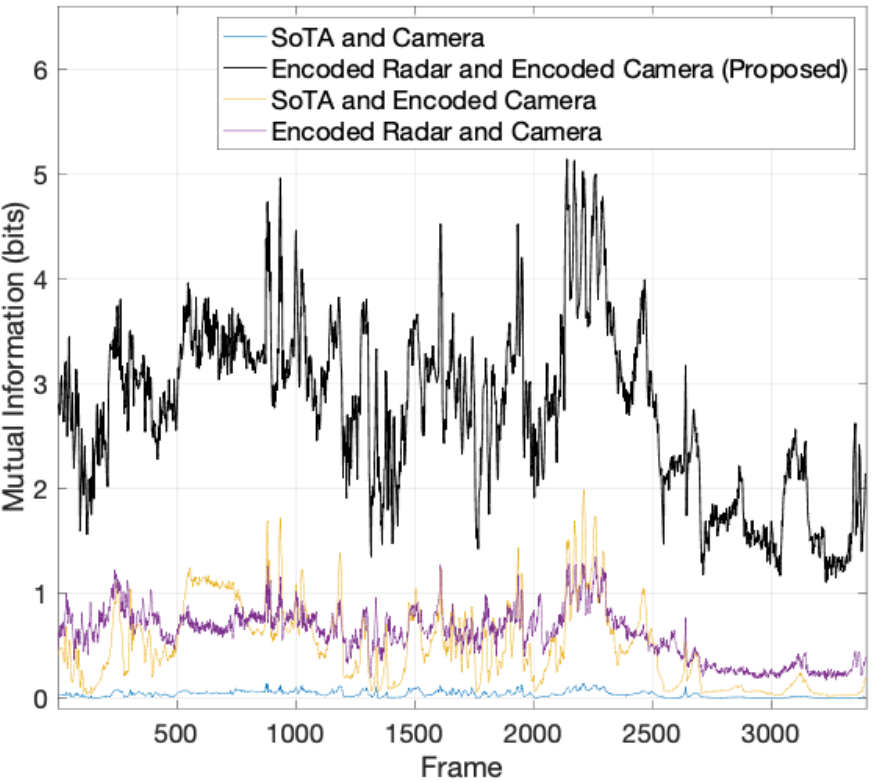}
		\caption{}
		\label{fig: Mutual Information}
	\end{subfigure}
        \hspace{0.01\textwidth}
	\begin{subfigure}{0.30\linewidth}
	        \includegraphics[width=\linewidth]{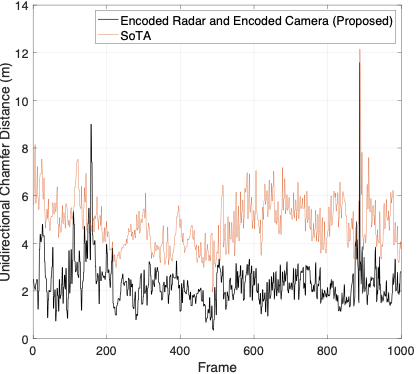}
	        \caption{}
	        \label{fig: UCD}
         \end{subfigure}
	\caption{(a) Correlation and (b) mutual information between several depth map pairs; and (c) UCD per frame. 'SoTA' refers to depth map obtained by \cite{roldan2024see} while 'Encoded' refers to the spectrum of encoded images using the described bases.}
	\label{fig:subfigures}
    \vspace{-14px}
\end{figure*}

\section{Experimental Results and Analysis}
We apply the proposed approach to the Radelft dataset's RGB images and radar depth maps of scenes 2 and 6, which include 5680 frames and are the only scenes that include SOTA 4D radar point clouds data. We performed two main experiments: a test of the spatial spectrum generation using the proposed approach and an enhancement of these spatial spectrum images using ResNet. 
\par We evaluate the performance of the data preprocessing by computing the Pearson correlation and mutual information metrics between the generated spectrums of RGB pixel semantic segmentation and the corresponding radar depth maps for $M_{radar}=20$ and $M_{cam}=200$, noting that a higher correlation value indicates a smaller discrepancy between the images, while mutual information indicates the learning potential of one modality from the other.

The performance of the depth map generation is measured by MAE, REL, UCD and BCD against the lidar point clouds and depth maps. MAE evaluates the average magnitude of errors in predictions, providing a straightforward measure of accuracy. REL normalizes the error by comparing it to the mean of the actual values, offering insight into the relative performance of the predictions. UCD measures the geometric similarity between the generated depth maps and the ground truth point clouds by calculating the average distance from each predicted point to its closest corresponding point in the lidar data. BCD extends upon UCD by measuring the average distance in both directions between the radar and lidar point sets.

\subsection{Data Preprocessing}
Our data preprocessing pipeline includes an input data conditioning submodule followed by the proposed encoding approach explained in the previous sections. We performed experiments for $M = 10, 20, 50, 70, 200$ and $\Phi=\Theta=(-70,70)$ degrees that truncate significant spectrum leakage at higher angles. The radar input is a simple data structure transformation from projected 3D point cloud coordinates to 2D depth maps, with the pixel value being inversely proportional to depth. We evaluate the preprocessing pipeline using Pearson correlation and mutual information. The higher value of Pearson correlation represents a stronger linear relationship between the two variables, while the higher value of mutual information indicates a greater reduction in entropy when predicting one variable from another.

\par Figure \ref{fig:algorithm_results} shows that larger values of $M$ result in higher-resolution images that better preserve object contours. The observed ripples along the horizontal and vertical axes are attributed to spectral leakage. Figure \ref{fig:Correlation} presents the frame-wise Pearson correlation between camera and radar modalities for $M_{radar}=20$ and $M_{cam}=200$. The results indicate that the proposed encoding significantly enhances the correlation when applied to both modalities. Notably, even single-modality encoding leads to a substantial increase in correlation. Similarly, Figure \ref{fig: Mutual Information} demonstrates that mutual information is significantly improved when both modalities are encoded, with considerable improvement also observed when encoding only one modality.

\par Tables \ref{tab:Correlation} and \ref{tab:Mutual Information} show the averages (mean values) of the plots in Figures \ref{fig:Correlation} and \ref{fig: Mutual Information}. The results show that there are improvements by factors of 3.88 and 76.69 in Pearson correlation and mutual information, respectively.
\vspace{-3px}

\begin{table}[h]
\renewcommand{\arraystretch}{1.2} 
\parbox{.45\linewidth}{
\centering
\begin{tabular}{|c||c c|}
\hline
    & $\textbf{I}_{cam}$ & $\textbf{P}_{cam}$ \\ \hline\hline
$\textbf{I}_{radar}$ & 0.1646 & 0.5503 \\
$\textbf{P}_{radar}$ & 0.0809 & \textbf{0.6396} \\ \hline
\end{tabular}
\caption{Average Pearson Correlation for different pairs.}
\label{tab:Correlation}
}
\hfill
\parbox{.45\linewidth}{
\centering
\begin{tabular}{|c||c c|}
\hline
    & $\textbf{I}_{cam}$ & $\textbf{P}_{cam}$ \\ \hline\hline
$\textbf{I}_{radar}$ & 0.0359 & 0.6117 \\ 
$\textbf{P}_{radar}$ & 0.4863 & \textbf{2.7533} \\ \hline
\end{tabular}
\caption{Average Mutual Information for different pairs.}
\label{tab:Mutual Information}
}
\vspace{-12px}
\end{table}

\begin{figure*}[t!]
  \centering
  \includegraphics[width=1\linewidth]{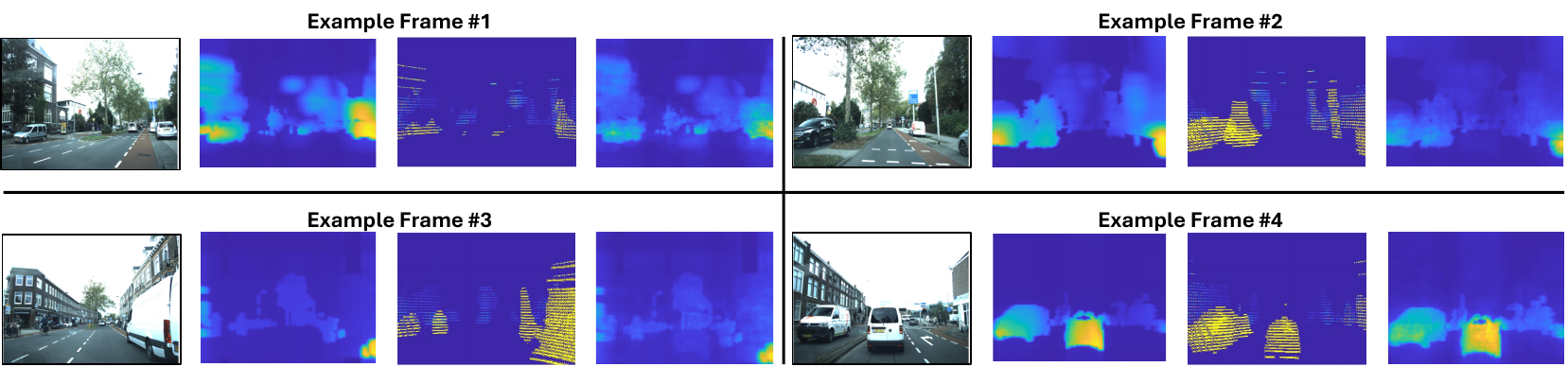}
  \caption{Results from the training module for four example frames. From left to right for each example frame: Scene in RGB, ground truth, original radar depth map, output from trained ResNet101. In each of the 4 frames, observe that our approach leads to sharper depth maps.}
  \label{fig:training_results}
  \centering
\end{figure*}

\begin{figure*}[!t]
	\centering
	\begin{subfigure}{0.24\linewidth}
		\includegraphics[width=\linewidth]{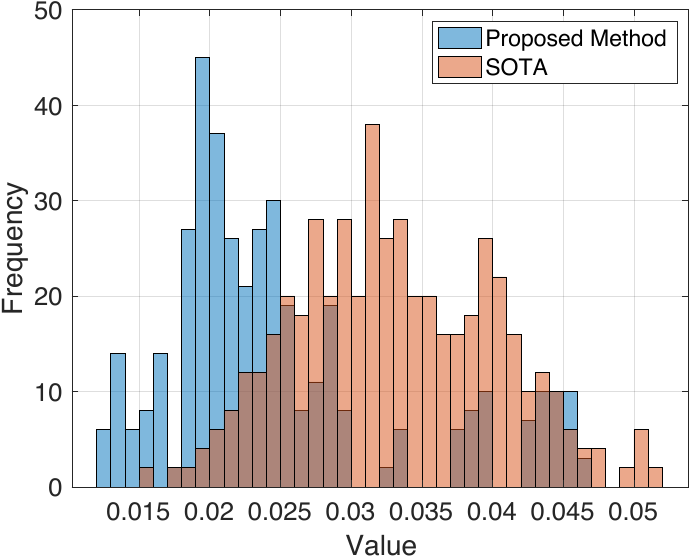}
		\caption{}
		\label{fig:hist_MAE}
	\end{subfigure}
	\begin{subfigure}{0.24\linewidth}
		\includegraphics[width=\linewidth]{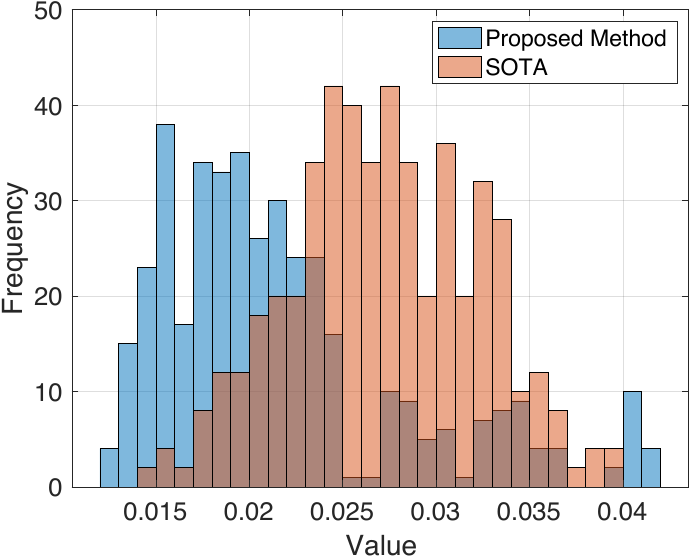}
		\caption{}
		\label{fig: hist_REL}
	\end{subfigure}
         \begin{subfigure}{0.24\linewidth}
	        \includegraphics[width=\linewidth]{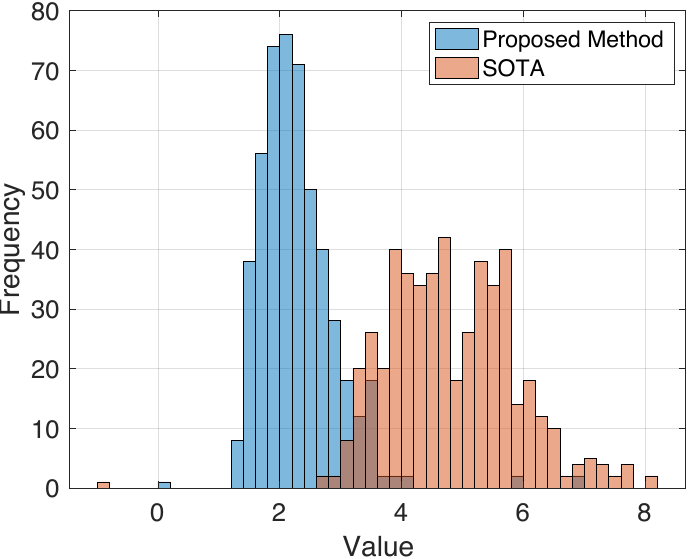}
	        \caption{}
	        \label{fig: hist_UCD}
         \end{subfigure}
         \begin{subfigure}{0.24\linewidth}
	        \includegraphics[width=\linewidth]{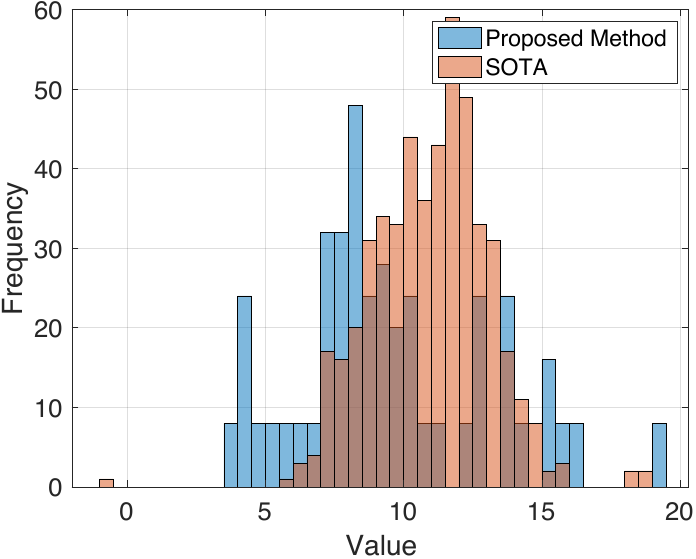}
	        \caption{}
	        \label{fig: hist_BCD}
         \end{subfigure}
	\caption{Error distributions: (a) MAE; (b) REL; (c) UCD; and (d) BCD.}
	\label{fig:histograms}
\end{figure*}

\subsection{Depth Map Generation}
The generation of depth maps is achieved with the ResNet101 network trained as described in the previous section with $M_{cam}=200$, $M_{radar}=20$ and $\Phi=\Theta=(-70,70)$. The ResNet101 is trained for 10,000 epochs, while the input data are compressed with the natural logarithm to make different features comparable in terms of scale. Qualitatively, from the results in Figure \ref{fig:training_results}, one can observe that the intensity locations in the resnet output are comparable to the locations in the ground truth images. In addition, the output depth maps show object contours clearly compared to the original radar depth map. However, insignificant impairments are visible because of the logarithmic feature compression and spectral leakage at the input.
\par Quantitatively, we evaluate the output using MAE, REL, UCD, and BCD. For UCD and BCD, the spectrum images are first transformed into 3D point clouds before the metrics are computed. Table \ref{tab:UCD} reports the results across different methods, including the SOTA deep neural network detector (SOTA DNN) \cite{roldan2024see} and OS-CFAR \cite{richards2010principles}. Figure \ref{fig: UCD} illustrates the per-frame UCD comparison between our method and the SOTA baseline. Our approach achieves improvements of 24.24\%, 18.52\%, 52.59\%, and 10.41\% in MAE, REL, UCD, and BCD, respectively.

\par Figure \ref{fig:histograms} presents the error histograms for each evaluation metric. For UCD, the proposed method demonstrate a clear improvement, with both the mean and standard deviation notably lower than those of the SOTA method. In the case of MAE and REL, the mean error is visibly reduced. Although the standard deviation is slightly higher, the increase is not substantially detrimental, given the significant reduction in mean value. For BCD, a lower mean is also observed compared to SOTA, but the increase in standard deviation is more pronounced than for the other metrics. Table \ref{tab: std} shows the standard deviation of each error metric. 

\renewcommand{\arraystretch}{1.4}
\begin{table}[h]
    \centering
    \begin{tabular}{|c|c|c|c|c|c|}
        \hline
        \textbf{Method} & \textbf{MAE} & \textbf{REL} & \textbf{UCD} & \textbf{BCD}\\
        \hline\hline
        Proposed Approach  & \textbf{0.025} & \textbf{0.022} & \textbf{2.29} & \textbf{9.81} \\
        SOTA DNN \cite{roldan2024see}         & 0.033 & 0.027 & 4.83 & 10.95 \\
        OS-CFAR \cite{richards2010principles}          & 0.029 & 0.152 & 18.02 & 232.29 \\
        \hline\hline
        \textbf{Improvement (\%)} & 24.24 & 18.52 & 52.59 & 10.41 \\
        \hline
    \end{tabular}
    \caption{Average MAE, REL, and UCD for different methods}
    \label{tab:UCD}
    \vspace{-7px}
\end{table}

\begin{table}[h]
    \centering
    \begin{tabular}{|c|c|c|c|c|c|}
        \hline
        \textbf{Method} & \textbf{MAE} & \textbf{REL} & \textbf{UCD} & \textbf{BCD}\\
        \hline\hline
        Proposed Approach  & 0.008 & 0.006 & 0.47 & 3.5 \\
        SOTA DNN \cite{roldan2024see}           & 0.007 & 0.005 & 1.05 & 2.51 \\
        \hline
    \end{tabular}
    \caption{Standard deviations of MAE, REL, and UCD for different methods}
    \label{tab: std}
    \vspace{-11px}
\end{table}

\begin{figure*}[!t]
	\centering
	\begin{subfigure}{0.31\linewidth}
		\includegraphics[width=\linewidth]{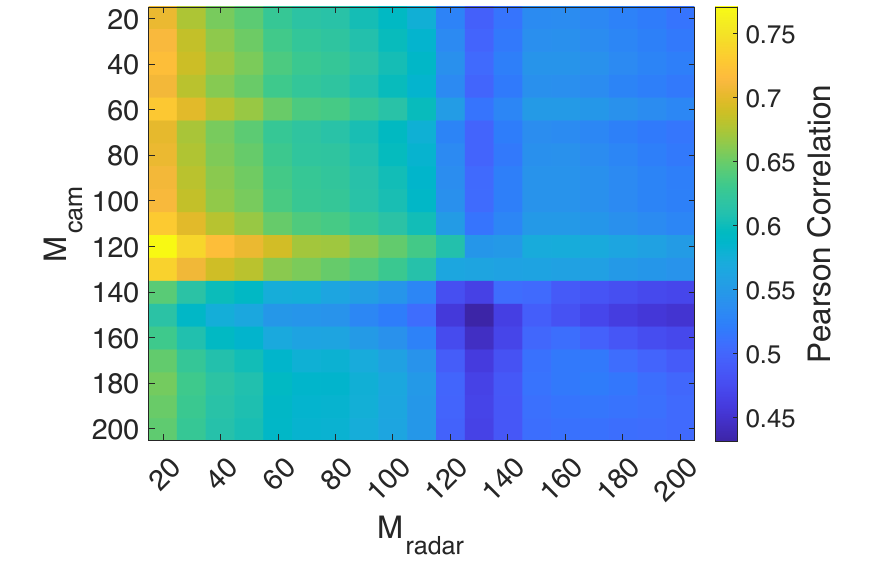}
		\caption{}
		\label{fig:ablation Correlation}
	\end{subfigure}
        \hspace{0.01\textwidth}
	\begin{subfigure}{0.31\linewidth}
		\includegraphics[width=\linewidth]{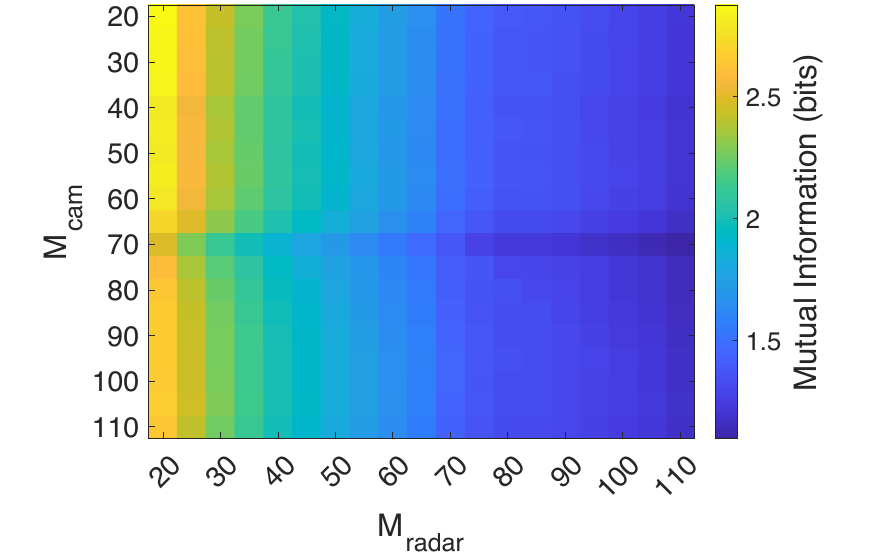}
		\caption{}
		\label{fig: ablation Mutual Information}
	\end{subfigure}
        \hspace{0.01\textwidth}
	\begin{subfigure}{0.31\linewidth}
	        \includegraphics[width=\linewidth]{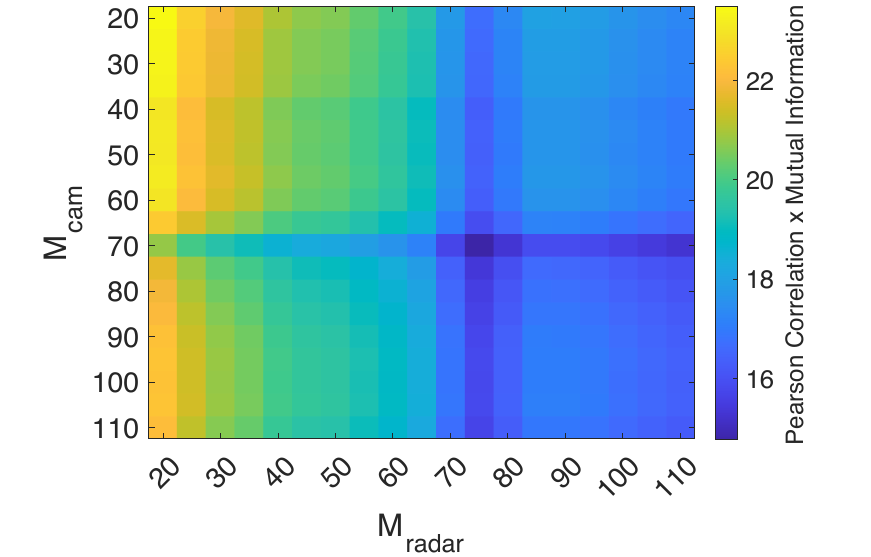}
	        \caption{}
	        \label{fig: multiplication}
         \end{subfigure}
	\caption{Results for different pairs of $M_{radar}$ and $M_{cam}$: (a) Pearson correlation coefficient; (b) mutual information; (c) element-wise multiplication of (a) and (b) showing best values for $M_{radar}$ and $M_{cam}$.}
	\label{fig: ablation SSIM}
\end{figure*}

\begin{figure*}[t!]
  \centering
  \includegraphics[width=\linewidth]{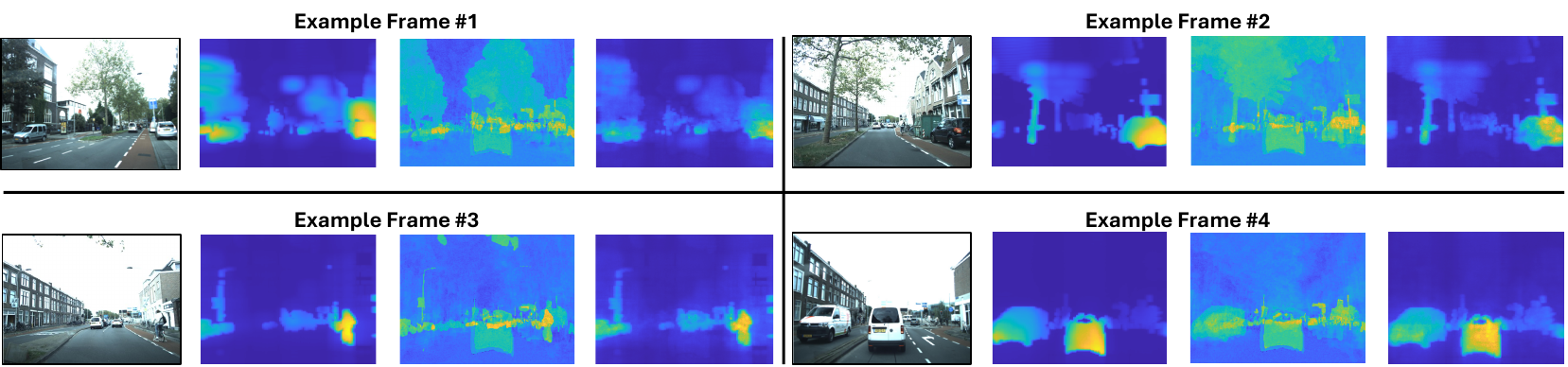}
  \caption{Results for ablation. Scene images left to right: camera RGB image, ground truth for training, spectrum estimation ablation result, and complete pipeline result.}
  \label{fig:ablation}
  \centering
  \vspace{-6px}
\end{figure*}

\begin{figure}[t!]
  \centering
  \includegraphics[width=0.7\linewidth]{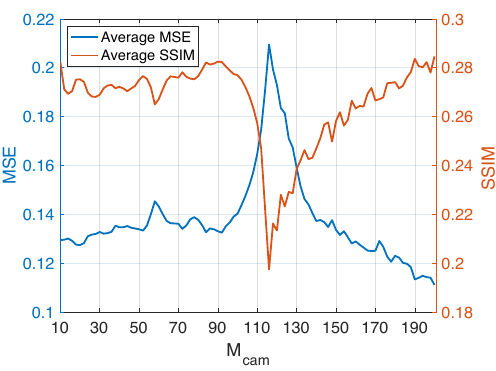}
  \caption{Average MSE and SSIM between RGB images semantic segmentation and its spectrum representation for different values of $M_{cam}$}
  \label{fig: MSE_LIST}
  \vspace{-11px}
\end{figure}

\section{Ablation Study}

To better understand the contribution of each component in our proposed framework, we conduct a comprehensive ablation study. This analysis isolates the effect of key design choices — including the number of segments, the quality of semantic segmentation, and the inclusion of spectrum representations — on the overall performance of the system. By experimenting with and modifying these components, we aim to find optimal settings for our pipeline.

\subsection{Number of segments}
As demonstrated in previous sections, the spectral representation of the radar depth map and the semantic segmentation of the RGB image enhance their mutual dependence, thus improving the learning process. As mentioned earlier, the resolution is directly proportional to \( M \), and a higher resolution spectrum for the radar depth maps is undesirable, since it may closely resemble the original depth maps rather than capturing the overall structure, as illustrated in Figure \ref{fig:algorithm_results}. In this section, we analyze the impact of different values of \( M \) selected from the set \( \{10, 12, 14, \dots, 200\} \), focusing on its effect on the cross-dependence between radar and camera, as well as its influence on preserving the overall structure in semantic segmentation.

\par Figure \ref{fig:ablation Correlation} shows the average Pearson correlation for different pairs of \( M_{cam} \) and \( M_{radar} \). It can be observed that the Pearson correlation is generally higher when \( M_{cam}, M_{radar} \leq 120 \), with a notable increase when \( M_{radar} \to 20 \). However, a high correlation is expected when both \( M_{cam} \) and \( M_{radar} \) approach low values (i.e., \( \leq 50 \)), as both semantic segmentation images and radar depth maps become blurry, thus capturing only the general structure of the scene without detailed contour information.

\par Figure \ref{fig: ablation Mutual Information} shows the mutual information value for different pairs of \( M_{cam} \) and \( M_{radar} \). Similarly to the Pearson correlation, it can be observed that mutual information increases as \( M_{radar} \) decreases. Furthermore, mutual information is maximized when both \( M_{cam} \) and \( M_{radar} \) approach 20. This is expected as contour information is not captured, which causes the semantic segmentation images to resemble blurry radar depth maps. As a result, both modalities become highly similar, with only the general structure of the scene being captured.

\par Figure \ref{fig: multiplication} shows the result of an element-wise multiplication of the Pearson correlation and mutual information maps, which yields the optimal values for \( M_{cam} \) and \( M_{radar} \). Based on the multiplication result, the best value for \( M_{radar} \) is 20, while \( M_{cam} \) can take any value except 120.

\par Figure~\ref{fig: MSE_LIST} illustrates the average mean squared error (MSE) and structural similarity index measure (SSIM) across multiple frames for each \( M_{\text{cam}} \). The average MSE gradually increases for \( M_{\text{cam}} \leq 90 \), followed by a significant rise within the range \( 100 \leq M_{\text{cam}} \leq 130 \), which we attribute to spectral leakage introduced by the estimation process. Beyond this range, the MSE decreases for \( M_{\text{cam}} \geq 130 \) and drops below 0.12 when \( M_{\text{cam}} \geq 180 \). The SSIM curve indicates that structural information is preserved for \( M_{\text{cam}} \leq 90 \) and reaches its maximum when \( M_{\text{cam}} \geq 180 \). In contrast, for \( 100 \leq M_{\text{cam}} \leq 130 \), the SSIM suggests that structural similarity is poorly maintained. Together, the MSE and SSIM curves suggest that best performance is achieved when \( M_{\text{cam}} \geq 180 \).

\subsection{Quality Semantic Segmentation}
Quality semantic segmentation is required to capture static and dynamic objects usually encountered on the road, such as poles, signs, trees, pedestrians, cars, cyclists, etc. Several semantic segmentation models were ready to use in our work, such as Deeplab (v1, v2 and v3) \cite{yurtkulu2019semantic}, Pyramid Scene Parsing Network (PSPNet) \cite{zhao2017pyramid}, and High-Resolution Network (HRNet) \cite{sun2019deep}. However, it should be noted that the choice of the semantic segmentation training benchmark is more important than the model itself. We found that the Cityscapes benchmark is optimal for capturing various objects, including objects captured by the 4D radar \cite{cordts2016cityscapes}. Other benchmarks like PASCAL (VOC) and ApolloScape were not sufficient alone to cover all objects present in the RaDelf dataset \cite{huang2018apolloscape, Everingham10}.

\subsection{Spectrum Representation}
Spectrum estimation serves as the final submodule in our pipeline. We qualitatively compare system performance with and without the spectrum estimation step. Figure \ref{fig:ablation} presents the qualitative results. It can be observed that the predicted depth map without spectral representations of the semantic segmentation and radar depth maps is less accurate compared to the depth map predicted with spectral representations. Specifically, the results indicate that ResNet struggles to accurately predict certain objects in the scene, leading to the generalization of some objects in the data set, as seen with the car in the example frame \#3. These findings highlight the importance of spectrum estimation in our framework.


\section{Conclusion and Future Work}

This paper presents a novel method for generating 4D radar depth maps for autonomous vehicles by integrating depth maps from DNN-based radar detectors with camera images through a positional encoding inspired by Bartlett's algorithm. The proposed approach transforms radar point clouds and camera images into a shared spectral representation, thereby enhancing cross-modal correspondence and enabling the training of radar image generators with high-resolution camera data. Experimental results demonstrate a 52.59\% improvement in UCD compared to SOTA, delivering more accurate and denser radar-generated depth maps than traditional CFAR detectors. By offering a highly trained 4D Radar, the presented approach holds the promise of making perception in AVs much more cost-effective when used in conjunction with cheap cameras in real time, by avoiding the expense of lidars. Although the mean BCD is significantly improved, future work may focus on reducing its standard deviation to achieve a more consistent error distribution.


\bibliographystyle{IEEEtran}
\bibliography{bibliography}

\begin{thebibliography}{10}
\providecommand{\url}[1]{#1}
\csname url@samestyle\endcsname
\providecommand{\newblock}{\relax}
\providecommand{\bibinfo}[2]{#2}
\providecommand{\BIBentrySTDinterwordspacing}{\spaceskip=0pt\relax}
\providecommand{\BIBentryALTinterwordstretchfactor}{4}
\providecommand{\BIBentryALTinterwordspacing}{\spaceskip=\fontdimen2\font plus
\BIBentryALTinterwordstretchfactor\fontdimen3\font minus \fontdimen4\font\relax}
\providecommand{\BIBforeignlanguage}[2]{{%
\expandafter\ifx\csname l@#1\endcsname\relax
\typeout{** WARNING: IEEEtran.bst: No hyphenation pattern has been}%
\typeout{** loaded for the language `#1'. Using the pattern for}%
\typeout{** the default language instead.}%
\else
\language=\csname l@#1\endcsname
\fi
#2}}
\providecommand{\BIBdecl}{\relax}
\BIBdecl

\bibitem{wang2019multi}
Z.~Wang, Y.~Wu, and Q.~Niu, ``Multi-sensor fusion in automated driving: A survey,'' \emph{Ieee Access}, vol.~8, pp. 2847--2868, 2019.

\bibitem{richards2010principles}
M.~A. Richards, J.~Scheer, W.~A. Holm, and W.~L. Melvin, ``Principles of modern radar,'' in \emph{Principles of Modern Radar}.\hskip 1em plus 0.5em minus 0.4em\relax Citeseer, 2010, ch.~16.

\bibitem{khan2022comprehensive}
M.~A.~U. Khan, D.~Nazir, A.~Pagani, H.~Mokayed, M.~Liwicki, D.~Stricker, and M.~Z. Afzal, ``A comprehensive survey of depth completion approaches,'' \emph{Sensors}, vol.~22, no.~18, p. 6969, 2022.

\bibitem{brodeski2019deep}
D.~Brodeski, I.~Bilik, and R.~Giryes, ``Deep radar detector,'' in \emph{2019 IEEE Radar Conference (RadarConf)}.\hskip 1em plus 0.5em minus 0.4em\relax IEEE, 2019, pp. 1--6.

\bibitem{cheng2022novel}
Y.~Cheng, J.~Su, M.~Jiang, and Y.~Liu, ``A novel radar point cloud generation method for robot environment perception,'' \emph{IEEE Transactions on Robotics}, vol.~38, no.~6, pp. 3754--3773, 2022.

\bibitem{roldan2024see}
I.~Roldan, A.~Palffy, J.~F. Kooij, D.~M. Gavrila, F.~Fioranelli, and A.~Yarovoy, ``See further than cfar: a data-driven radar detector trained by lidar,'' \emph{arXiv preprint arXiv:2402.12970}, 2024.

\bibitem{chen2017rethinking}
L.-C. Chen, G.~Papandreou, F.~Schroff, and H.~Adam, ``Rethinking atrous convolution for semantic image segmentation,'' \emph{arXiv preprint arXiv:1706.05587}, 2017.

\bibitem{Bartlett1948}
M.~S. Bartlett, ``Smoothing periodograms from time-series with continuous spectra,'' \emph{Nature}, vol. 161, no. 4096, pp. 686--687, 1948.

\bibitem{gao2019deep}
J.~Gao, X.~Yi, C.~Zhong, X.~Chen, and Z.~Zhang, ``Deep learning for spectrum sensing,'' \emph{IEEE Wireless Communications Letters}, vol.~8, no.~6, pp. 1727--1730, 2019.

\bibitem{VoD}
A.~Palffy, E.~Pool, S.~Baratam, J.~F.~P. Kooij, and D.~M. Gavrila, ``Multi-class road user detection with 3+1d radar in the view-of-delft dataset,'' \emph{IEEE Robotics and Automation Letters}, vol.~7, no.~2, pp. 4961--4968, 2022.

\bibitem{kradar}
D.-H. Paek, S.-H. Kong, and K.~T. Wijaya, ``K-radar: 4d radar object detection for autonomous driving in various weather conditions,'' \emph{Advances in Neural Information Processing Systems}, vol.~35, pp. 3819--3829, 2022.

\bibitem{long2021radar}
Y.~Long, D.~Morris, X.~Liu, M.~Castro, P.~Chakravarty, and P.~Narayanan, ``Radar-camera pixel depth association for depth completion,'' in \emph{Proceedings of the IEEE/CVF Conference on Computer Vision and Pattern Recognition}, 2021, pp. 12\,507--12\,516.

\bibitem{singh2023depth}
A.~D. Singh, Y.~Ba, A.~Sarker, H.~Zhang, A.~Kadambi, S.~Soatto, M.~Srivastava, and A.~Wong, ``Depth estimation from camera image and mmwave radar point cloud,'' in \emph{Proceedings of the IEEE/CVF Conference on Computer Vision and Pattern Recognition}, 2023, pp. 9275--9285.

\bibitem{li2024radarcam}
H.~Li, Y.~Ma, Y.~Gu, K.~Hu, Y.~Liu, and X.~Zuo, ``Radarcam-depth: Radar-camera fusion for depth estimation with learned metric scale,'' \emph{arXiv preprint arXiv:2401.04325}, 2024.

\bibitem{li2024semantic}
Z.~Li, Y.~Song, F.~Ai, C.~Song, and Z.~Xu, ``Semantic-guided depth completion from monocular images and 4d radar data,'' \emph{IEEE Transactions on Intelligent Vehicles}, 2024.

\bibitem{radarocc}
F.~Ding, X.~Wen, Y.~Zhu, Y.~Li, and C.~X. Lu, ``Radarocc: Robust 3d occupancy prediction with 4d imaging radar,'' \emph{arXiv preprint arXiv:2405.14014}, 2024.

\bibitem{AoA}
A.~Dixit and et~al., ``Detection and localization of targets using millimeter wave radars: An experimental study,'' in \emph{2021 IEEE International Conference on Electronics, Computing and Communication Technologies (CONECCT)}, 2021, pp. 1--6.

\bibitem{priestley1981spectral}
M.~B. Priestley, \emph{Spectral analysis and time series}.\hskip 1em plus 0.5em minus 0.4em\relax Academic press London, 1981, vol. 890.

\bibitem{xu2024overview}
Z.-Q.~J. Xu, Y.~Zhang, and T.~Luo, ``Overview frequency principle/spectral bias in deep learning,'' \emph{Communications on Applied Mathematics and Computation}, pp. 1--38, 2024.

\bibitem{pillai2012array}
S.~U. Pillai, \emph{Array signal processing}.\hskip 1em plus 0.5em minus 0.4em\relax Springer Science \& Business Media, 2012.

\bibitem{buckley1990spatial}
K.~M. Buckley and X.-L. Xu, ``Spatial-spectrum estimation in a location sector,'' \emph{IEEE transactions on acoustics, speech, and signal processing}, vol.~38, no.~11, pp. 1842--1852, 1990.

\bibitem{yurtkulu2019semantic}
S.~C. Yurtkulu, Y.~H. {\c{S}}ahin, and G.~Unal, ``Semantic segmentation with extended deeplabv3 architecture,'' in \emph{2019 27th Signal Processing and Communications Applications Conference (SIU)}.\hskip 1em plus 0.5em minus 0.4em\relax IEEE, 2019, pp. 1--4.

\bibitem{cordts2016cityscapes}
M.~Cordts, M.~Omran, S.~Ramos, T.~Rehfeld, M.~Enzweiler, R.~Benenson, U.~Franke, S.~Roth, and B.~Schiele, ``The cityscapes dataset for semantic urban scene understanding,'' in \emph{Proceedings of the IEEE conference on computer vision and pattern recognition}, 2016, pp. 3213--3223.

\bibitem{zhao2017pyramid}
H.~Zhao, J.~Shi, X.~Qi, X.~Wang, and J.~Jia, ``Pyramid scene parsing network,'' in \emph{Proceedings of the IEEE conference on computer vision and pattern recognition}, 2017, pp. 2881--2890.

\bibitem{sun2019deep}
K.~Sun, B.~Xiao, D.~Liu, and J.~Wang, ``Deep high-resolution representation learning for human pose estimation,'' in \emph{Proceedings of the IEEE/CVF conference on computer vision and pattern recognition}, 2019, pp. 5693--5703.

\bibitem{huang2018apolloscape}
X.~Huang, X.~Cheng, Q.~Geng, B.~Cao, D.~Zhou, P.~Wang, Y.~Lin, and R.~Yang, ``The apolloscape dataset for autonomous driving,'' in \emph{Proceedings of the IEEE conference on computer vision and pattern recognition workshops}, 2018, pp. 954--960.

\bibitem{Everingham10}
M.~Everingham and et~al., ``The pascal visual object classes (voc) challenge,'' \emph{International Journal of Computer Vision}, vol.~88, no.~2, pp. 303--338, Jun. 2010.

\end{thebibliography}

\end{document}